# Integrating E-Commerce and Data Mining: Architecture and Challenges

Suhail Ansari, Ron Kohavi, Llew Mason, and Zijian Zheng

Blue Martini Software
2600 Campus Drive
San Mateo, CA, 94403, USA

{suhail,ronnyk,lmason,zijian}@bluemartini.com

## Abstract

We show that the e-commerce domain can provide all the right ingredients for successful data mining and claim that it is a killer domain for data mining. We describe an integrated architecture, based on our experience at Blue Martini Software, for supporting this integration. The architecture can dramatically reduce the pre-processing, cleaning, and data understanding effort often documented to take 80% of the time in knowledge discovery projects. We emphasize the need for data collection at the application server layer (not the web server) in order to support logging of data and metadata that is essential to the discovery process. We describe the data transformation bridges required from the transaction processing systems and customer event streams (e.g., clickstreams) to the data warehouse. We detail the mining workbench, which needs to provide multiple views of the data through reporting, data mining algorithms, visualization, and OLAP. We conclude with a set of challenges.

## 1 Introduction

E-commerce is growing fast, and with this growth companies are willing to spend more on improving the online experience. In *Commerce Software Takes Off* [1], the authors from Forrester Research wrote that online business to consumer retail spending in 1999 was $20.3 billion and estimated to grow to $144 billion by 2003. Global 2500 companies will spend 72% more on e-commerce in 2000 than they did in 1999. Existing sites are using primitive measures, such as page views, but the need for more serious analysis and personalization is growing quickly with the need to differentiate. In *Measuring Web Success* [2], the authors claim that "Leaders will use metrics to fuel personalization" and that "firms need web intelligence, not log analysis."

Data Mining tools aid the discovery of patterns in data.[1] Until recently, companies that have concentrated on building horizontal data mining modeling tools, have had little commercial success. Many companies were bought, including the acquisition of Compression Sciences by Gentia for $3 million, HyperParallel by Yahoo for about $2.3 million, Clementine by SPSS for $7 million, and Thinking Machines's Darwin by Oracle for less than $25 million. Recently, a phase shift has occurred in the valuation of such companies, and recent acquisitions have given rise to valuations 10 to 100 times higher. KD1 was acquired by Net Perceptions for $116M, RightPoint (previously DataMind) was acquired by E.piphany for $400M, DataSage was acquired by Vignette for $577M, and NeoVista was acquired by Accrue for $140M. The shift in valuations indicates wider recognition of the value of data mining modeling techniques for e-commerce.

E-commerce is the killer-domain for data mining. It is ideal because many of the ingredients required for successful data mining are easily satisfied: data records are plentiful, electronic collection provides reliable data, insight can easily be turned into action, and return on investment can be measured. To really take advan-tage of this domain, however, data mining must be integrated into the e-commerce systems with the appropriate data transformation bridges from the transaction processing system to the data warehouse and vice-versa. Such integration can dramatically reduce the data preparation time, known to take about 80% of the time to complete an analysis [3]. An integrated solution can also provide users with a uniform user interface and seamless access to metadata.

---

[1] In this paper, we use the term *data mining* to denote the wider process, sometimes called *knowledge discovery*, which includes multiple disciplines, such as preprocessing, reporting, exploratory analysis, visualization, and modeling.

The paper is organized as follows. Section 2 describes the integrated architecture that we propose, explaining the main components and the bridges connecting them. Section 3 details the data collector, which must collect much more data than what is available using web server log files. Section 4 describes the analysis component, which must provide a breadth of data transformation facilities and analysis tools. We describe a set of challenging problems in Section 5, and conclude with a summary in Section 6.

## 2 Integrated Architecture

In this section we give a high level overview of a proposed architecture for an e-commerce system with intregrated data mining. Details of the most important parts of the architecture and their advantages appear in following sections. The described system is an ideal architecture based on our experiences at Blue Martini Software. However, we make no claim that everything described here is implemented in Blue Martini Software's products. In our proposed architecture there are three main components, *Business Data Definition*, *Customer Interaction*, and *Analysis*. Connecting these components are three data transfer bridges, *Stage Data*, *Build Data Warehouse,* and *Deploy Results*. The relationship between the components and the data transfer bridges is illustrated in Figure 1. Next we describe each component in the architecture and then the bridges that connect these components.

In the *Business Data Definition* component the e-commerce business user defines the data and metadata associated with their business. This data includes merchandising information (e.g., products, assortments, and price lists), content information (e.g., web page templates, articles, images, and multimedia) and business rules (e.g., personalized content rules, promotion rules, and rules for cross-sells and up-sells). From a data mining perspective the key to the *Business Data Definition* component is the ability to define a rich set of attributes (metadata) for any type of data. For example, products can have attributes like size, color, and targeted age group, and can be arranged in a hierarchy representing categories like men's and women's, and subcategories like shoes and shirts. As another example, web page templates can have attributes indicating whether they show products, search results, or are used as part of the checkout process. Having a diverse set of available attributes is not only essential for data mining, but also for personalizing the customer experience.

The *Customer Interaction* component provides the interface between customers and the e-commerce business. Although we use the example of a web site throughout this paper, the term customer interaction applies more generally to any sort of interaction with customers. This interaction could take place through a web site (e.g., a marketing site or a web store), customer service (via telephony or email), wireless application, or even a bricks-and-mortar point of sale system. For effective analysis of all of these data sources, a data collector needs to be an integrated part of the *Customer Interaction* component. To provide maximum utility, the data collector should not only log sale transactions, but it should also log other types of cus-tomer interactions, such as web page views for a web site. Further details of the data collection architecture for the specific case of a web site are described in Section 3. To illustrate the utility of this integrated data collection let us consider the example of an e-commerce company measuring the effectiveness of its web banner advertisements on other sites geared at attracting customers to its own site. A similar analysis can be applied when measuring the effectiveness of advertising or different personalizations on its own site.

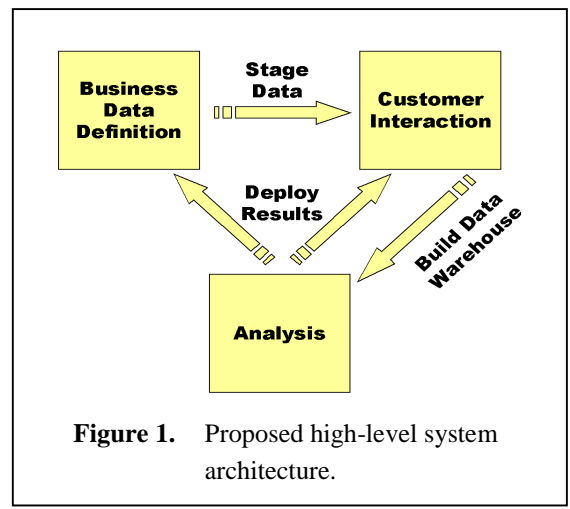

**Figure 1.** Proposed high-level system architecture.

The cost of a web banner advertisement is typically based on the number of "click-throughs." That is, there is a fee paid for each visitor who clicks on the banner advertisement. Many e-commerce companies measure the *effectiveness* of their web banner advertisements using the same metric, the number of click-throughs, and thus fail to take into account the *sales generated* by each referred visitor. If the goal is to sell more products then the site needs to attract buyers rather than browsers. A recent Forrester Research report [2] stated that *"Using hits and page views to judge site success is like evaluating a musical performance by its volume."* In practice, we have seen the ratio of generated sales to click-throughs vary by as much as a factor of 20 across a company's web banner advertisements. One advertisement generated five times as much in sales as another advertisement, even though clickthroughs from the former advertisement were one quarter of the clickstreams from the latter. The ability to measure this sort of relationship requires conflation of multiple data sources.

The *Analysis* component provides an integrated environment for decision support utilizing data transformations, reporting, data mining algorithms, visualization, and OLAP tools. The richness of the available metadata gives the *Analysis* component significant advantages over horizontal decision support tools, in both power and ease-of-use. For instance, the system automatically knows the type of each attribute, including whether a discrete attribute's values are ordered, whether the range of a continuous attribute is bounded, and textual descriptions. For a web site, the system knows that each customer has web sessions and that each web session includes page views and orders. This makes it a simple matter to compute aggregate statistics for combinations of customers, sessions, page views, and orders automatically. We examine the integrated analysis component in more detail in Section 4.

The *Stage Data* bridge connects the *Business Data Definition* component to the *Customer Interaction* component. This bridge transfers (or *stages*) the data and metadata into the *Customer Interaction* component. Having a staging process has several advantages, including the ability to test changes before having them implemented in production, allowing for changes in the data formats and replication between the two components for efficiency, and enabling e-commerce businesses to have zero down-time.

The *Build Data Warehouse* bridge links the *Customer Interaction* component with the *Analysis* component. This bridge transfers the data collected within the *Customer Interaction* component to the *Analysis* component and builds a data warehouse for analysis purposes. The *Build Data Warehouse* bridge also transfers all of the business data defined within the *Business Data Definition* component (which was transferred to the *Customer Interaction* component using the *Stage Data* bridge). The data collector in the *Customer Interaction* component is usually implemented within an On-Line Transaction Processing (OLTP) system, typically designed using entity relation modeling techniques. OLTP systems are geared towards efficient handling of a large number of small updates and short queries. This is critical for running an e-commerce business, but is not appropriate for analysis [4, 5], which usually requires full scans of several very large tables and a star schema design which business users can understand. For data mining, we need to build a data warehouse using dimensional modeling techniques. Both the data warehouse design and the data transfer from the OLTP system to the data warehouse system are very complex and time-consuming tasks. Making the construction of the data warehouse an integral part of the architecture significantly reduces the complexity of these tasks. In addition to typical ETL (Extract, Transform and Load) functionality, the bridge supports import and integration of data from both external systems and syndicated data providers (e.g.,

Acxiom). Since the schema in the OLTP system is controlled by the architecture, we can automatically convert the OLTP schema to a multi-dimensional star schema that is optimized for analysis.

The last bridge, *Deploy Results*, is the key to "closing the loop" and making analytical results actionable. It provides the ability to transfer models, scores, results and new attributes constructed using data transformations back into the *Business Data Definition* and *Customer Interaction* components for use in business rules for personalization. For example, customers can be scored on their propensity to accept a cross-sell and the site can be personalized based on these scores. This is arguably the most difficult part of the knowledge discovery process to implement in a non-integrated system. However, the shared metadata across all three components means that results can be directly reflected in the data which defines the e-commerce company's business.

## 3 Data Collection

This section describes the data collection component of the proposed architecture. This component logs customers' transactions (e.g., purchases and returns) and event streams (e.g., clickstreams). While the data collection component is a part of every customer touch point (e.g., web site, customer service applications, and wireless applications), in this section we will describe in detail the data collection at the web site. Most of the concepts and techniques mentioned in this section could be easily extended to other customer touch points.

### 3.1 Clickstream Logging

Most e-commerce architectures rely on web server logs or packet sniffers as a source for clickstream data. While both these systems have the advantage of being non-intrusive, allowing them to "bolt on" to any e-commerce application, they fall short in logging high level events and lack the capability to exploit metadata available in the application. A typical web log contains data such as the page requested, time of request, client HTTP address, etc., for each web server request. For each page that is requested from the web server, there are a huge number of requests for images and other content on the page. Since all of these are recorded in the web server logs, most of the data in the logs relates to requests for image files that are mostly useless for analysis and are commonly filtered out. All these requests need to be purged from the web logs before they can be used. Because of the stateless nature of HTTP, each request in a web log appears independent of other requests, so it becomes extremely difficult to identify users and user sessions from this data [6, 7, 8, 9]. Since the web logs only contain the name of the page that was requested, these page names have to be mapped to the content, products, etc., on the page. This problem is further compounded by the introduction of dynamic content where the same page can be used to display different content for each user. In this case, details of the content displayed on a web page may not even be captured in the web log. The mechanism used to send request data to the server also affects the information in the web logs. If the browser sends a request using the "POST" method, then the input parameters for this request are not recorded in the web log.

Packet sniffers try to collect similar data by looking at data "on the wire." While packet sniffers can "see" more data than what is present in web logs, they still have problems identifying users (e.g., same visitor logging in from two different machines) and sessions. Also, given the myriad ways in which web sites are designed it is extremely difficult to extract logical business information by looking at data streaming across a wire. To further complicate things, packet sniffers can't see the data in areas of the site that are encoded for secure transmission and thus have difficulty working with sites (or areas of a site) that use SSL (Secure Socket Layer). Such areas of a site are the most crucial for analysis including checkout and forms containing personal data. In many financial sites including banks, the entire site is secure thus making packet sniffers that monitor the encrypted data blind and essentially useless, so the sniffers must be given access to data prior to encryption, which complicates their integration.

Collecting data at the application server layer can effectively solve all these problems. Since the application server serves the content (e.g., images, products and articles), it has detailed knowledge of the content being served. This is true even when the content is dynamically generated or encoded for transmission using SSL. Application servers use cookies (or URL encoding in the absence of cookies) to keep track of a user's session, so "sessionizing" the clickstream is trivial. Since the application server also keeps track of the user, using login mechanisms or cookies, associating the clickstream with a particular visitor is simple. The application server can also be designed to keep track of information absent in web server logs including pages that were aborted (user pressed the "stop" button while the page was being downloaded), local time of the user, speed of the user's connection and if the user had turned their cookies off. This method of collecting clickstream data has significant advantages over both web logs and packet sniffers.

## 3.2 Business Event Logging

The clickstream data collected from the application server is rich and interesting; however, significant insight can be gained by looking at subsets of requests as one logical event or episode [6, 10]. We call these aggregations of requests *business events*. Business events can also be used to describe significant user actions like sending an email or searching [2]. Since the application server has to maintain the context of a user's session and related data, the application server is the logical choice for logging these business events. Business events can be used to track things like the contents of abandoned shopping carts, which are extremely difficult to track using only clickstream data. Business events also enable marketers to look beyond page hit rates to *micro-conversion rates* [11]. A micro-conversion rate is defined for each step of the purchasing process as the fraction of products that are successfully carried through to the next step of the purchasing process. Two examples of these are the fraction of product views that resulted in the product being added to the shopping cart and the fraction of products in the shopping cart that successfully passed through each phase of the checkout process. Thus the integrated approach proposed in this architecture gives marketers the ability to look directly at product views, content views, and product sales, a capability far more powerful than just page views and click-throughs. Some interesting business events that help with the analysis given above and are supported by the architecture are

- Add/Remove item to/from shopping cart
- Initiate checkout
- Finish checkout
- Search event
- Register event

The search keywords and the number of results for each of these searches that can be logged with the search events give marketers significant insight into the interests of their visitors and the effectiveness of the search mechanism.

## 3.3 Measuring Personalization Success

The architecture also supports a rules engine that runs on the web site for personalization. Rules can be deployed for offering promotions to visitors, displaying specific products or content to a specific visitor, etc. After the rules are deployed, business events can be used to track the effect of deploying these rules. A business event can be collected each time that a rule is used in personalization and these events, coupled with the shopping-cart/checkout events, can give an excellent estimate of the effectiveness of each rule. The architecture can also use control groups so that personalization rules are only activated for a fraction of the target visitors. This enables analysts to directly look at sales or results for visitors when the rules were and were not activated.

Similar data collection techniques can be used for all the customer touch points like customer service representatives, wireless applications, etc. Collecting the right data is critical to effective analysis of an e-commerce operation.

## 4 Analysis

This section describes the analysis component of our architecture. We start with a discussion of data transformations, followed by analysis techniques including reporting, data mining algorithms, visualiza-tion, and OLAP. The data warehouse is the source data of analyses in our architecture. Although dimensional modeling is usually a prerequisite for analysis, our experience shows that many analyses require additional data transformations that convert the data into forms more amenable to data mining.

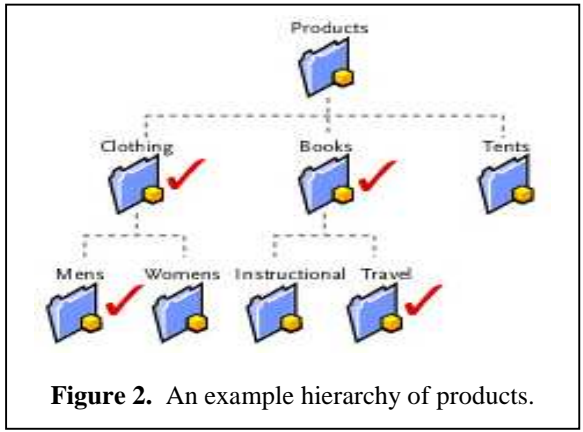

**Figure 2.** An example hierarchy of products.

As we mentioned earlier, the business user can define product, promotion, and assortment hierarchies in the *Business Data Definition* component. Figure 2 gives a simple example of a product hierarchy. This hierarchical information is very valuable for analysis, but few existing data mining algorithms can utilize it directly. Therefore, we need data transformations to convert this information to a format that can be used by data mining algorithms. One possible solution is to add a column indicating whether the item falls under a given node of the hierarchy. Let us use the product hierarchy shown in Figure 2 as an example. For each order line or page request containing a product SKU (Stock Keeping Unit), this transformation creates a Boolean column corresponding to each selected node in the hierarchy. It indicates whether this product SKU belongs to the product category represented by the node. Figure 3 shows the enriched row from this operation.

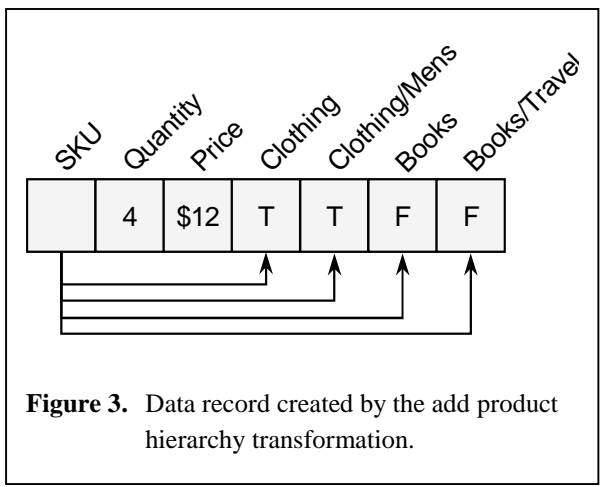

**Figure 3.** Data record created by the add product hierarchy transformation.

Since customers are the main concern of any e-commerce business, most data mining analyses are at the customer level. That is, each record of a data set at the final stage of an analysis is a customer signature containing all the information about the customer. However, the majority of the data in the data warehouse is at other levels such as the order header level, the order line level, and the page request level. Each customer may have multiple rows at these levels. To make this detailed information useful for analyses at the customer level, aggregation transformations are necessary. Here are some examples of attributes we have found useful:

- What percentage of each customer's orders used a VISA credit card?
- How much money does each customer spend on books?
- How much is each customer's average order amount above the mean value of the average order amount for female customers?
- What is the total amount of each customer's five most recent purchases over $30?
- What is the frequency of each customer's purchases?
- What is the recency of each customer's purchases (the number of days since the last purchase)?

These attributes are very hard to construct using standard SQL statements, and need powerful aggregation transformations. We have found RFM (Recency, Frequency, and Monetary) attributes particularly useful for the e-commerce domain.

E-commerce data contains many date and time columns. We have found that these date and time columns convey useful information that can reveal im-portant patterns. However, the common date and time format containing the year, month, day, hour, minute, and second is not often supported by data mining algorithms. Most patterns involving date and time cannot be directly discovered from this format. To make the discovery of patterns involving dates and times easier, we need transformations which can compute the time difference between dates (e.g., order date and ship date), and create new attributes representing day-of-week, day-of-month, week, month, quarter, year, etc. from date and time attributes.

Based on the considerations mentioned above, the architecture is designed to support a rich set of transformations. We have found that transformations including: create new attributes, add hierarchy attributes, aggregate, filter, sample, delete columns, and score are useful for making analyses easier.

With transformations described, let us discuss the analysis tools. Basic reporting is a bare necessity for e-commerce. Through generated reports, business users can understand how a web site is working at different levels and from different points of view. Example questions that can be answered using reporting are:
- What are the top selling products?
- What are the worst selling products?
- What are the top viewed pages?
- What are the top failed searches?
- What are the conversion rates by brand?
- What is the distribution of web browsers?
- What are the top referrers by visit count?
- What are the top referrers by sales amount?
- What are the top abandoned products?

Our experience shows that some reporting questions such as the last two mentioned above are very hard to answer without an integrated architecture that records both event streams and sales data.

Model generation using data mining algorithms is a key component of the architecture. It reveals patterns about customers, their purchases, page views, etc. By generating models, we can answer questions like:
- What characterizes heavy spenders?
- What characterizes customers that prefer promotion X over Y?
- What characterizes customers that accept cross-sells and up-sells?
- What characterizes customers that buy quickly?
- What characterizes visitors that do not buy?

Based on our experience, in addition to automatic data mining algorithms, it is necessary to provide interactive model modification tools to support business insight. Models either automatically generated or created by interactive modifications can then be examined or evaluated on test data. The purpose is to let business users understand their models before deploying them. For example, we have found that for rule models, measures such as confidence, lift, and support at the individual rule level and the individual conjunct level are very useful in addition to the overall accuracy of the model. In our experience, the following functionality is useful for interactively modifying a rule model:
- Being able to view the segment (e.g., customer segments) defined by a subset of rules or a subset of conjuncts of a rule.
- Being able to manually modify a rule model by deleting, adding, or changing a rule or individual conjunct.

For example, a rule model predicting heavy spenders contains the rule:

```
IF    Income > $80,000 AND
      Age <= 31 AND
      Average Session Duration is between
         10 and 20.1 minutes AND
      Account creation date is before
         2000-04-01
THEN  Heavy spender
```

It is very likely that you wonder why the split on age occurs at 31 instead of 30 and the split on average session duration occurs at 20.1 minutes instead of 20 minutes. Why does account creation date appear in the rule at all? A business user may want to change the rule to:

```
IF    Income > $80,000 AND
      Age <= 30 AND
      Average Session Duration is between
          10 and 20 minutes
THEN  Heavy spender
```

However, before doing so, it is important to see how this changes the measures (e.g. confidence, lift, and support) of this rule and the whole rule model.

Given that humans are very good at identifying patterns from visualized data, visualization and OLAP tools can greatly help business users to gain insight into business problems by complementing reporting tools and data mining algorithms. Our experience suggests that visualization tools are very helpful in understanding generated models, web site operations, and data itself. Figure 4 shows an example of a visualization tool, which clearly reveals that females aged between 30 and 39 years are heavy spenders (large square), closely followed by males aged between 40 and 49 years.

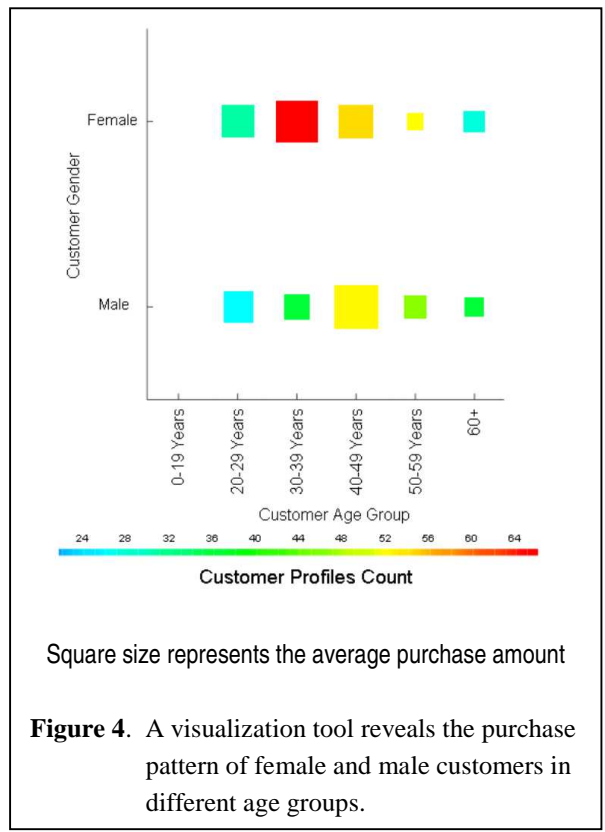

Square size represents the average purchase amount

**Figure 4**. A visualization tool reveals the purchase pattern of female and male customers in different age groups.

## 5  Challenges

In this section we describe several challenging problems based on our experiences in mining e-commerce data. The complexity and granularity of these problems differ, but each represents a real-life area where we believe improvements can be made. Except for the first two challenges, the problems deal with data mining algorithmic challenges.

**Make Data Mining Models Comprehensible to Business Users**

Business users, from merchandisers who make decisions about the assortments of products to creative designers who design web sites to marketers who decide where to spend advertising dollars, need to understand the results of data mining. Summary reports are easiest to understand and usually easy to provide, especially for specific vertical domains. Simple visualizations, such as bar charts and two-dimensional scatterplots, are also easy to understand and can provide more information and highlight patterns, especially if used in conjunction with color. Few data mining models, however, are easy to understand. Classification rules are the easiest, followed by classification trees. A visualization for the Naïve-Bayes classifier [12] was also easy for business users to understand in the second author's past experience.

The challenge is to define more model types (hypothesis spaces) and ways of presenting them to business users. What regression models can we come up with and how can we present them? (Even linear regression is usually hard for business users to understand.) How can we present nearest-neighbor models, for example? How can we present the results of association rule algorithms without overwhelming users with tens of thousands of rules (a nice example of this problem can be found in Berry and Linoff [13] starting on page 426)?

**Make Data Transformation and Model Building Accessible to Business Users**

The ability to answer a question given by a business user usually requires some data transformations and technical understanding of the tools. Our experience is that even commercial report designers and OLAP tools are too hard for most business users. Two common solutions are (i) provide templates (e.g., reporting templates, OLAP cubes, and recommended transformations for mining) for common questions, something that works well in well-defined vertical markets, and (ii) provide the expertise through consulting or a services organization. The challenge is to find ways to empower business users so that they will be able to serve themselves.

**Support Multiple Granularity Levels**

Data collected in a typical web site contains records at different levels of granularity:
- Page views are the lowest level with attributes such as product viewed and duration.
- Sessions include attributes such as browser used, initiation time, referring site, and cookie information. Each session includes multiple page views.
- Customer attributes include name, address, and demographic attributes. Each customer may be involved in multiple sessions.

Mining at the page view level by joining all the session and customer attributes violates the basic assumption inherent in most data mining algorithms, namely that records are independently and identically distributed. If we are trying to build a model to predict who visits page X, and Joe happens to visit it very often, then we might get a rule that if the visitor's first name is Joe, they will likely visit page X. The rule will have multiple records (visits) to support it, but it clearly will not generalize beyond the specific Joe. This problem is shared by mining problems in the telecommunication domain [14]. The challenge is to design algorithms that can support multiple granularity levels correctly.

**Utilize Hierarchies**

Products are commonly organized in hierarchies: SKUs are derived from products, which are derived from product families, which are derived from categories, etc. A product hierarchy is usually three to eight levels deep. A customer purchases SKU level items, but generalizations are likely to be found at higher levels (e.g., families and categories). Some algorithms have been designed to support tree-structured attributes [15], but they do not scale to the large product hierarchies. The challenge is to support such hierarchies within the data mining algorithms.

**Scale Better: Handle Large Amounts of Data**

Yahoo! had 465 million page views per day in December of 1999 [16]. The challenge is to find useful techniques (other than sampling) that will scale to this volume of data. Are there aggregations that should be performed on the fly as data is collected?

**Support and Model External Events**

External events, such as marketing campaigns (e.g., promotions and media ads), and site redesigns change patterns in the data. The challenge is to be able to model such events, which create new patterns that spike and decay over time.

**Support Slowly Changing Dimensions**

Visitors' demographics change: people get married, their children grow, their salaries change, etc. With these changes, their needs, which are being modeled, change. Product attributes change: new choices (e.g., colors) may be available, packaging material or design change, and even quality may improve or degrade. These attributes that change over time are often referred to as "slowly changing dimensions" [4]. The challenge is to keep track of these changes and provide support for such changes in the analyses.

**Identify Bots and Crawlers**

Bots and crawlers can dramatically change click-stream patterns at a web site. For example, Keynote (www.keynote.com) provides site performance measurements. The Keynote bot can generate a request multiple times a minute, 24 hours a day, 7 days a week, skewing the statistics about the number of sessions, page hits, and exit pages (last page at each session). Search engines conduct breadth first scans of the site, generating many requests in short duration. Internet Explorer 5.0 supports automatic synchronization of web pages when a user logs in, when the computer is idle, or on a specified schedule; it also supports offline browsing, which loads pages to a specified depth from a given page. These options create additional clickstreams and patterns. Identifying such bots to filter their clickstreams is a non-trivial task, especially for bots that pretend to be real users.

## 6  Summary

We proposed an architecture that successfully integrates data mining with an e-commerce system. The proposed architecture consists of three main components: *Business Data Definition*, *Customer Interaction*, and *Analysis*, which are connected using data transfer bridges. This integration effectively solves several major problems associated with horizontal data mining tools including the enormous effort required in pre-processing of the data before it can be used for mining, and making the results of mining actionable. The tight integration between the three components of the architecture allows for automated construction of a data warehouse within the *Analysis* component. The shared metadata across the three components further simplifies this construction, and, coupled with the rich set of mining algorithms and analysis tools (like visualization, reporting and OLAP) also increases the efficiency of the knowledge discovery process. The tight integration and shared metadata also make it easy to deploy results, effectively closing the loop. Finally we presented several challenging problems that need to be addressed for further enhancement of this architecture.

## Acknowledgments

We would like to thank other members of the data mining and visualization teams at Blue Martini Software and our documentation writer, Cindy Hall. We wish to thank our clients for sharing their data with us and helping us refine our architecture and improve Blue Martini's products.